\documentclass[runningheads]{llncs}

\usepackage{hyperref}
\usepackage{subcaption}
\usepackage{setspace}
\usepackage{placeins}
\usepackage{color}
\usepackage{amsmath}
\usepackage{graphicx}
\usepackage{amssymb}

\begin{document}

\title{$\mathsf{Safe PILCO}$: a software tool\\ for safe and data-efficient policy synthesis}

\author{Kyriakos Polymenakos \and
Nikitas Rontsis \and\\
Alessandro Abate \and
Stephen Roberts}
\institute{University of Oxford, Oxford, UK \\
\email{kpol@robots.ox.ac.uk}}

\authorrunning{Polymenakos et al.}

\maketitle

\begin{abstract}
$\mathsf{Safe PILCO}$ is a software tool for safe and data-efficient policy search with reinforcement learning. 
It extends the known $\mathsf{PILCO}$ algorithm, originally written in MATLAB, to support safe learning.  
We provide a Python implementation  and leverage existing libraries that allow the codebase to remain short and modular,
which is appropriate for wider use by the verification, reinforcement learning, and control communities. 
\end{abstract}


\section{Goals and design philosophy}
\sloppy
Reinforcement learning (RL) is a well-known, widely-used framework that has recently enjoyed breakthroughs using  model-free methods based on deep neural networks \cite{dqn,duan2016benchmarking,haarnoja2018soft}. Notable shortcomings of model-free deep RL algorithms are their need for extensive training datasets, the lack of interpretability, and the difficulty to verify their outcomes. 
It is data-efficient, which makes it appealing for applications involving physical systems. 
$\mathsf{PILCO}$ \cite{pilco} (Probabilistic Inference for Learning COntrol) represents a state-of-the-art model-based RL method that relies on Gaussian processes (thus, not on deep neural networks). 
So far, $\mathsf{PILCO}$ does not incorporate safety constraints 
and comes as a MATLAB implementation.  
$\mathsf{SafePILCO}$, based on \cite{polymenakos2019safe}, extends the original algorithm with 
safety constraints embedded in the training procedure and as learning goals, 
and comes as a concise, clean and efficient Python implementation. 



$\mathsf{SafePILCO}$ is underpinned by an object-oriented architecture, enabling code re-use by keeping the implementation short and modular, with the capability to flexibly replace individual components.  
It takes advantage of available open source libraries, both as building blocks of the core algorithm, and as predefined tasks for evaluating the performance of the algorithm. It uses standard libraries to implement specific sub-tasks and to facilitate extensions, e.g. the GPflow library give access to an array of models with a consistent interface. Additionally, by using standard scenarios for experimental evaluation, $\mathsf{SafePILCO}$ enables users to employ it as a benchmark to easily compare their own methods against. 


\section{Description of the Software Tool} \label{sec:description}

$\mathsf{SafePILCO}$ comes as an open source Python package\footnote{Main package repository: \url{https://github.com/nrontsis/PILCO}}.
To make reproduction of the experiments easier we provide additional functionalities (such as logging, post processing results and creating the plots in the paper) in a separate repository\footnote{Experiments and figures reproduction repository: \url{https://github.com/kyr-pol/SafePILCO_Tool-Reproducibility}}.
In a standard object-oriented fashion, the main components of the algorithm are organised as objects, following a hierarchy of classes. The main components are:
\begin{itemize}
\item the Gaussian process model, providing short-term and long-term predictions; 
\item the policy or controller, which selects an action based on the state observation at every time step; 
\item the parametric reward function, which captures the performance of the algorithm and is also tasked with enforcing safe behaviour; 
\item scenarios that capture environment dynamics specific to a case study. 
\end{itemize}

Firstly, the environment the agent interacts with needs to be specified.
$\mathsf{SafePILCO}$ is designed to seamlessly interface with any environment following the OpenAI gym API. 
Therefore, gym environments can be directly invoked, 
as well as user-defined environments equipped with the necessary functionalities. 

\begin{figure}
    \centering
    \includegraphics[width=0.7\linewidth]{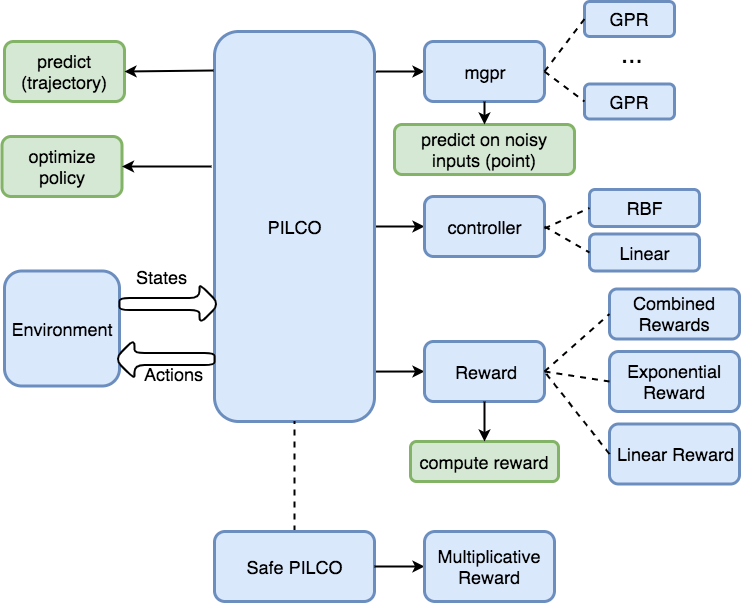}
    \caption{ \small The basic structure of the $\mathsf{SafePILCO}$ implementation. Black arrows correspond to object-attribute relationship, dashed lines to inheritance, and wide arrows to data flow. Classes are represented by blue boxes and key functions by green boxes.}
    \label{fig:struct}
\vspace{-0.4cm}
\end{figure}

\sloppy
The \verb|PILCO| class is the central object of the package, 
encapsulating the GP model, the controller, and the reward function as attributes. 
\verb|PILCO| employs the model and the controller to predict a trajectory, 
calls the reward function to evaluate it, 
and uses the gradients calculated through automatic differentiation in combination with an external optimiser, to improve the controller parameters. 
The \verb|SafePILCO| subclass combines the common additive reward function component used for performance, with a multiplicative component that encodes the risk of violating the safety requirement over any time step of the episode (this is used to enforce safety during training).  

The \verb|mgpr| class implements the multi-input, multi-output Gaussian process regression that underpins the dynamical model. Specifically, \verb|mgpr| combines several, multi-input/single-output GP models. These GP models are provided by GPflow, along with standard GP inference and prediction. Our code newly provides GP predictions for multiple output dimensions, when 
the inputs are multi-dimensional and noisy. 
The \verb|mgpr| class also allows
defining priors for the GP hyperparameters.

The policy (or the controller) defines how the agent selects appropriate actions at each time step.
Policies are implemented as memory-less, deterministic feedback controllers: 
the control input is thus directly dependent on the environment state that the agent observes at the current time step. The agent implements a policy $\pi$ of the form $u = \pi^{\theta}(x)$, 
where $\theta$ are the policy parameters. 
The package provides the \verb|controller| class with two subclasses, one for \emph{linear} controllers, and one for controllers based on \emph{radial basis functions} (RBF). The only extra requirement from the controller is the ability to calculate, for 
a Gaussian-distributed state (including the predicted states during the planning phase), a similarly
Gaussian-distributed control input, so that the state and input are \emph{jointly} Gaussian.
The policies are parametric and optimising the values of these parameters $\theta$ constitutes the overall policy search objective. 
 
The final part concerns the specification of the reward function.  
We note that this is somewhat different from most of the RL literature: in $\mathsf{SafePILCO}$, much like for the original $\mathsf{PILCO}$ \cite{pilco} algorithm, the reward function is known analytically \textit{a-priori}. 
The \verb|reward| class implements the standard reward, while also encoding an adaptively weighted penalty that encourages constraint satisfaction. 
Having an analytic expression for the reward function is necessary for the GP model to estimate the reward of a proposed policy, without interacting with the environment. The class provides, at any given state, scalar outputs that capture the expected reward and the constraint violation probability. A composite reward function that is used to train the policy thus has one component that evaluates the performance of the agent and one that evaluates the safety of the policy.  
We further note that it is the choice of reward function, along with the environment, that defines a task: indeed, we can design multiple tasks with a shared environment by varying the reward function. This setup is therefore suitable for transfer learning, or for multi-task learning \cite{deisenroth2014multi}, since the same model is valid across multiple tasks. 

\subsection{Libraries}
The tool relies on other Python packages, allowing us to leverage their optimised functionalities and to keep the codebase succinct. 
Furthermore, this allows users to easily apply our algorithm to new tasks. 
We use Tensorflow \cite{tensorflow2015-whitepaper} to obtain automatic gradient computations (often referred to as auto-diff), which thus simplifies the policy improvement step.\footnote{By way of comparison, all gradient calculations in the $\mathsf{PILCO}$ Matlab implementation are hand-coded, thus extensions are laborious as any additional user-defined controller or reward function has to include these gradient calculations too.} 
GPflow \cite{GPflow2017} is a Python package for Gaussian Process modelling built on Tensorflow. GPflow provides a full set of basic GP functionalities, and gives access to many specialised models. 
Having a Tensorflow back-end, gradients in all the GPflow models are also calculated automatically. 
Additionally, GPflow allows the user to readily define priors and to employ different optimisers or alternative implementations of sparse approximations for GPs. Finally, our implementation is interfaced with the Open-AI gym \cite{openai}, a suite of RL tasks widely used in the community.  
Gym tasks have consistent interfaces and detailed visualisation capabilities for a wide range of tasks varying over different sorts of complexity: dimensionality, smoothness of dynamics, length of episodes, and so on. 
Users can quickly prototype their algorithms using easier tasks and move to more complex, time-consuming experiments, as the project matures.



\newlength{\subfigWidth}
\setlength{\subfigWidth}{0.17\textwidth}
\begin{figure*}[t]
        \centering
        \begin{subfigure}[t]{\subfigWidth}
            \centering
            \includegraphics[width=\textwidth]{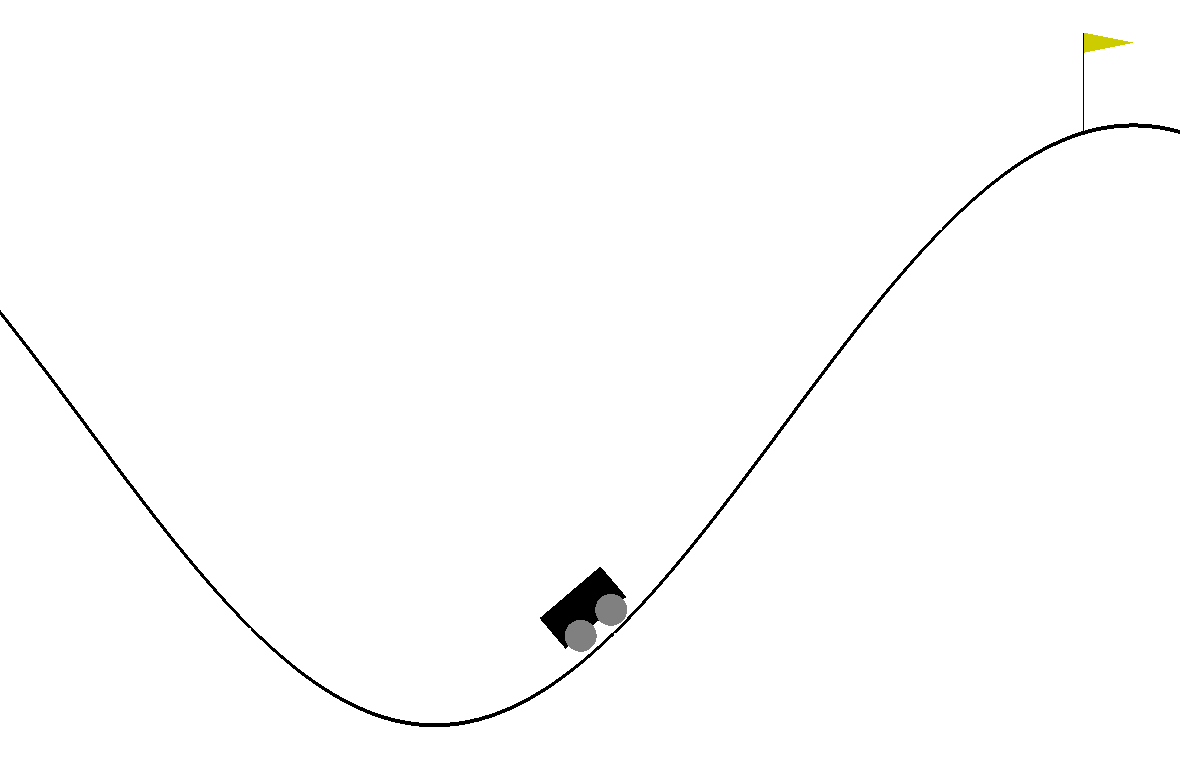}
            \caption{\small Mountain Car}    
        \end{subfigure}
        \quad
        \begin{subfigure}[t]{\subfigWidth}  
            \centering 
            \includegraphics[width=\textwidth, height=0.80\textwidth]{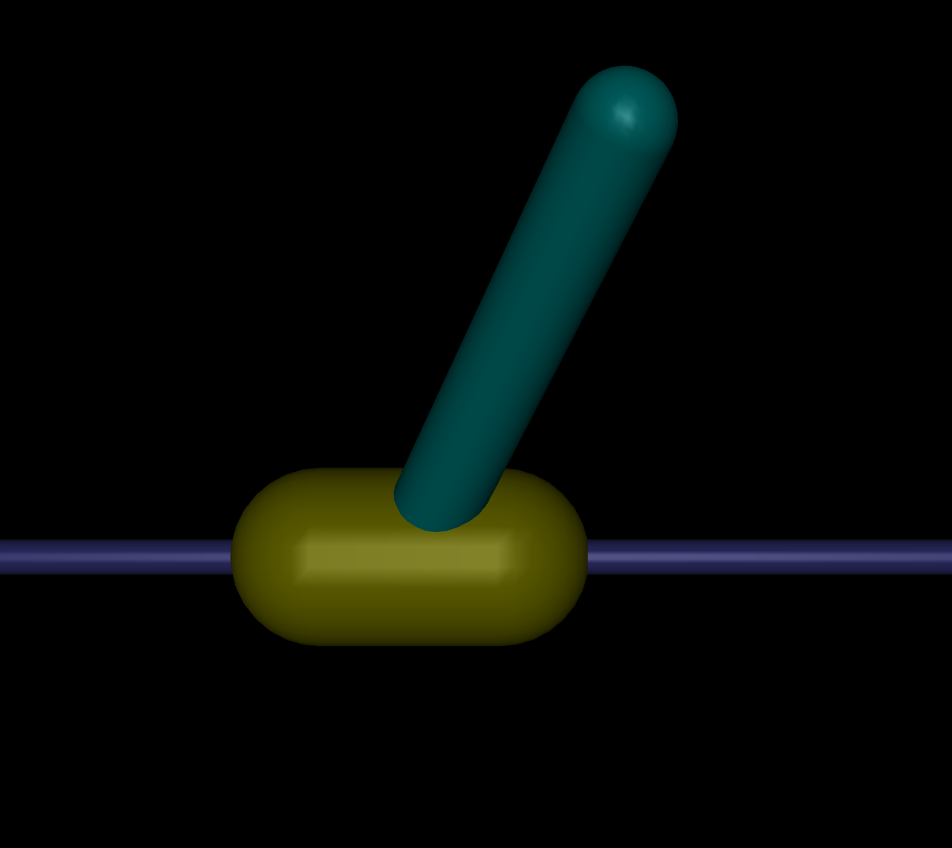}
            \caption{\small Inverted Pendulum}    
        \end{subfigure}
        \quad
        \begin{subfigure}[t]{\subfigWidth}   
            \centering 
            \includegraphics[width=\textwidth, height=0.80\textwidth]{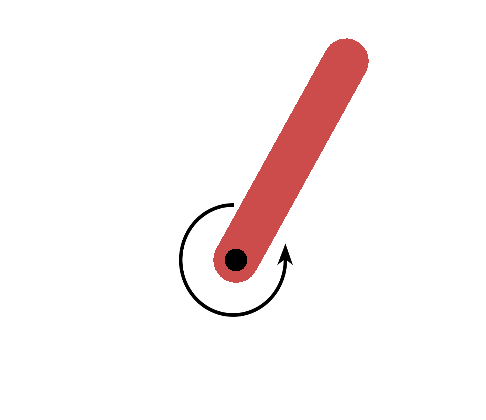}
            \caption{\small Pendulum swing-up}
        \end{subfigure}
        \quad
        \begin{subfigure}[t]{\subfigWidth}   
            \centering 
            \includegraphics[width=\textwidth,height=0.80\textwidth]{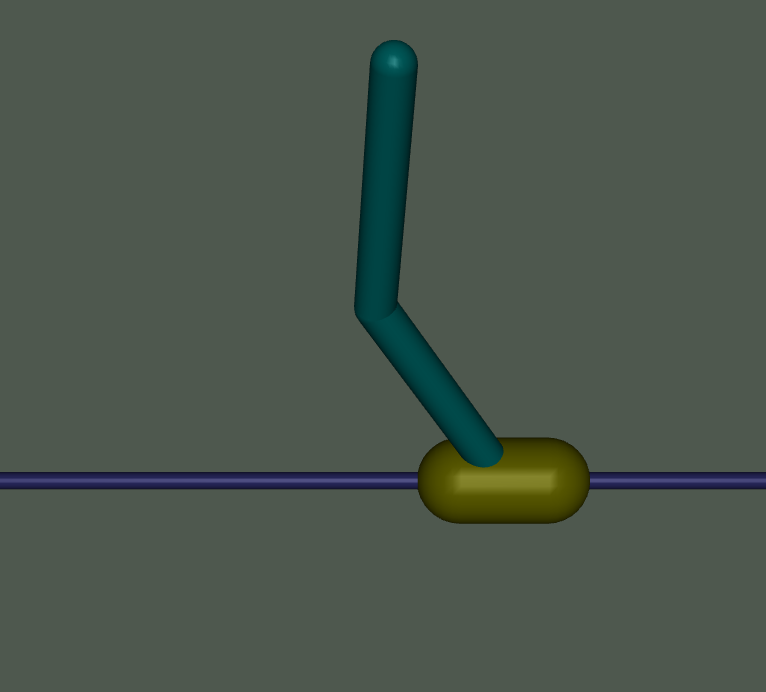}
            \caption{\small Double pendulum}    
        \end{subfigure}
        \quad
        \begin{subfigure}[t]{\subfigWidth}   
            \centering 
            \includegraphics[width=\textwidth, height=0.80\textwidth]{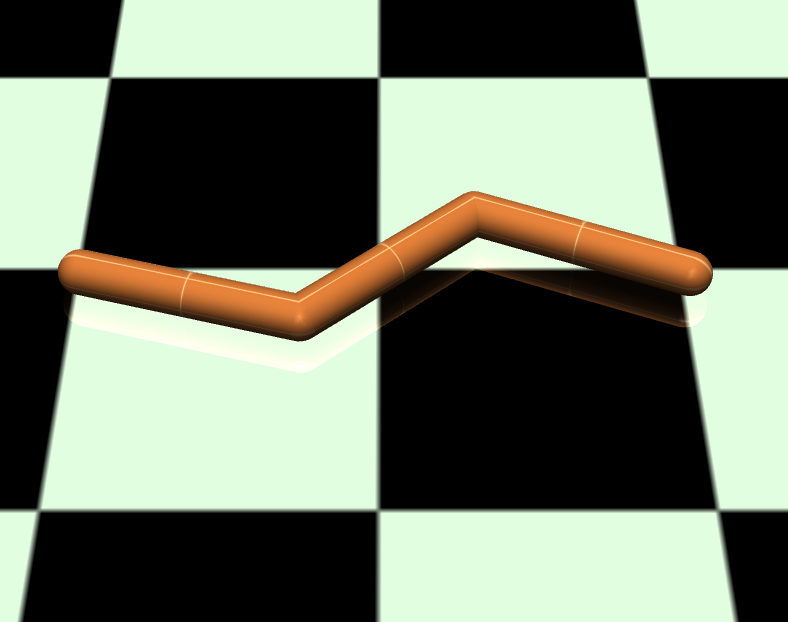}
            \caption{\small Swimmer}
            \label{fig:swimmer}
        \end{subfigure}
        \caption{\small Snapshots of the Open-AI gym environments used in the case studies.}
        \label{fig:env_screenshots}
\end{figure*}

\section{Case Studies}

To evaluate the performance of the package we run a set of experiments on different tasks. The results reported are averaged over 10 random seeds, along with standard deviations. To obtain a more accurate evaluation of the controller at each iteration, for each random seed, we test it 5 times and take the mean (variance is not used). Visualisations of the environments used for the case studies are shown in Figure \ref{fig:env_screenshots}. 

Details of the OpenAI gym tasks that are used with no modification are in \cite{openai} or on the gym website\footnote{\url{https://gym.openai.com/} \label{gym}}. Notations and key sourcecode variables are presented in Table \ref{table:hypers}. All hyperparameters used and key environment characteristics are summarised in Table \ref{table:hypers2}. 
Experiments are presented in order of increasing complexity. 
As mentioned previously in Section \ref{sec:description},  $\mathsf{SafePILCO}$ assumes a predetermined, closed-form reward function.
Most of the tasks we apply our method on come with their own reward functions, that do not follow the analytical form that is assumed. 
Thus we make the following distinction:
the algorithm is evaluated on the original reward function coming with the environments, but is trained with a closed-from reward function of our design. 
Designing a reward function that leads to a desired behaviour (in our case, behaviour that maximises accumulated return measured with a different reward function), is thoroughly studied and can prove challenging \cite{ng2003shaping,mataric1994reward}, 
but in our experience and for the environments used, it was not too involved and did not require extensive hyperparameter searches or particularly careful manual tuning.

\begin{figure}[t]
        \centering
        \begin{subfigure}[t]{0.30\textwidth}
            \centering
            \includegraphics[width=\textwidth]{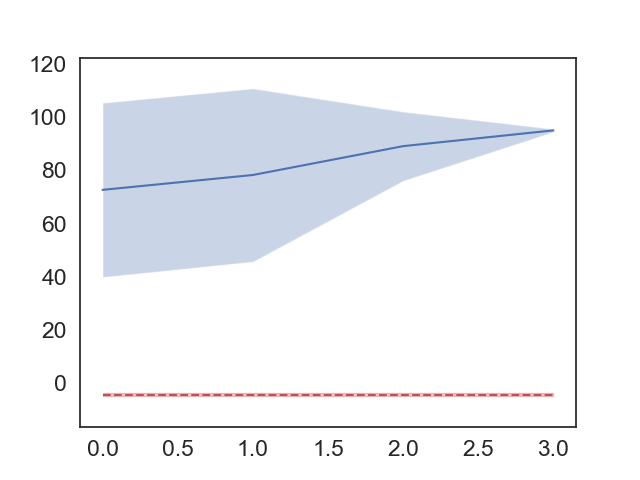}
            \caption{\small Mountain Car}    
        \end{subfigure}
        \quad
        \begin{subfigure}[t]{0.30\textwidth}  
            \centering 
            \includegraphics[width=\textwidth]{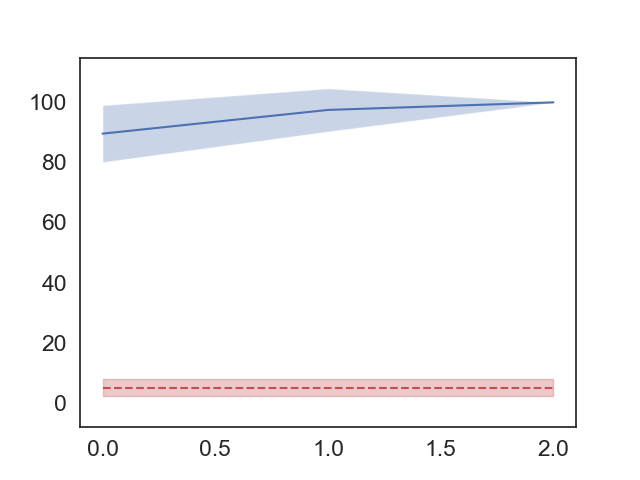}
            \caption{\small Inverted Pendulum}    
        \end{subfigure}
        \quad
        \begin{subfigure}[t]{0.30\textwidth}   
            \centering 
            \includegraphics[width=\textwidth]{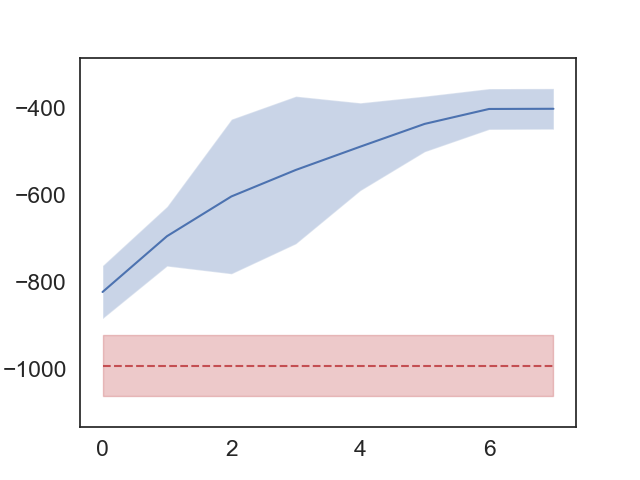}
            \caption{\small Pendulum swing-up}
        \end{subfigure}
        \vskip\baselineskip
        \begin{subfigure}[t]{0.30\textwidth}   
            \centering 
            \includegraphics[width=\textwidth]{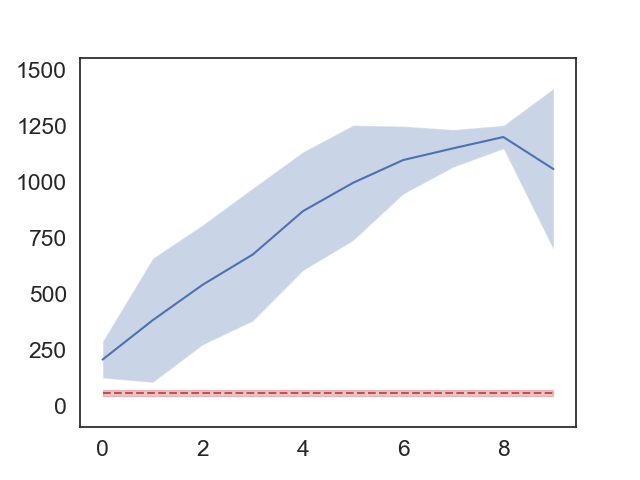}
            \caption{\small Double pendulum}
        \end{subfigure}
        \quad
        \begin{subfigure}[t]{0.30\textwidth}   
            \centering 
            \includegraphics[width=\textwidth]{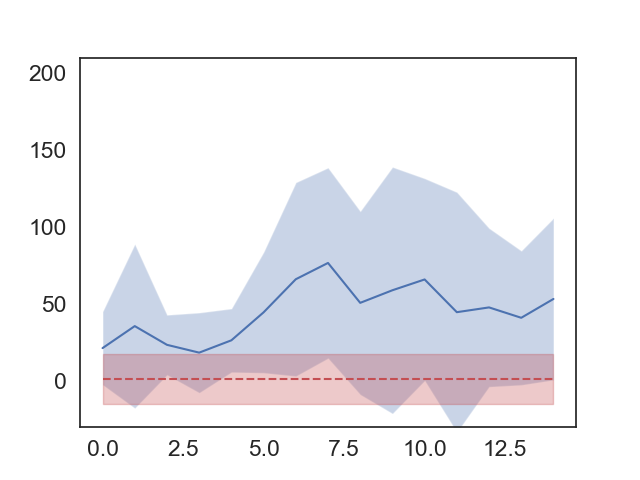}
            \caption{\small Swimmer, average performance}    
        \end{subfigure}
        \quad
        \begin{subfigure}[t]{0.30\textwidth}   
            \centering 
            \includegraphics[width=\textwidth]{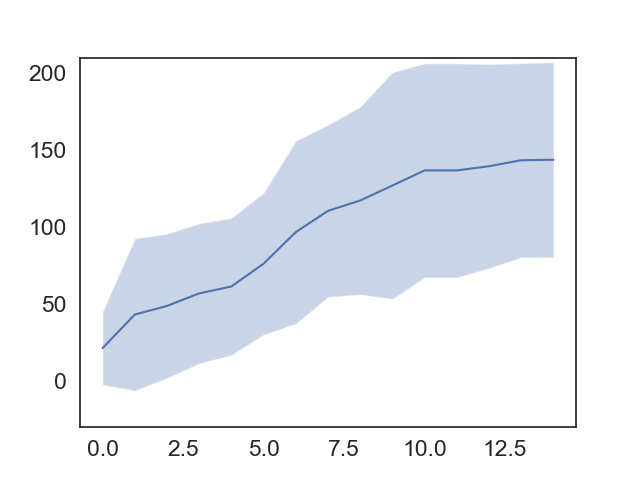}
            \caption{\small Swimmer, best performance so far}
        \end{subfigure}
        \caption
        {\small Experimental results for different OpenAI gym tasks. Episode rewards on the y-axis and iteration number on the x-axis   
        (blue, mean and two standard deviations around it).
        The performance of a random policy (red dashed line) is shown for comparison.
        For the swimmer we report both the average performance of 10 random seeds at each iteration, 
        and the best performance so far in all previous iterations of each random seed.}
        \label{fig:plainpilco}
        \vspace{-0.1cm}
    \end{figure}

\subsection{Plain PILCO} \label{sec:plain}

We give specific information for each environment and the associated reward functions. 
The mountain car experiment uses the \verb|MountainCarContinuous-v0| gym environment. 
Small negative rewards are given at those states where the car is not at the top of the hill (goal state), whereas a large reward is given exclusively as the agent gets to the top. The goal state is the terminal state for the environment and no further reward is obtained. This is easily captured with a negative exponential reward, centered at the goal state.

Next, the OpenAI gym \cite{openai} \verb|InvertedPendulum-v2| environment is used. It is a variant of the cart-pole
stabilisation task, where a pendulum is attached to a cart on a rail, and the controller applies a force to the cart. 
The pendulum starts close the upright position and the controller's task is to stabilise it by moving the cart to the left or to the right on the rail. 
For this scenario, as well as for the double inverted pendulum scenario (see below), the native reward function provides a $+1$ reward when the pendulum angle is less than some given threshold. Once out of this area the episode terminates, as the controller cannot exert a stabilising input.  An exponential reward centered at the upright position is again used. 

For the pendulum swing up task, we modify part of the default behaviour of the gym \verb|Pendulum-v0| environment: 
the initial starting state distribution of the pendulum is too wide 
for a feasible unimodal planning. 
We restrict this initial distribution to the pendulum starting close to the downward position.
The environment penalises an agent with negative rewards correlated to the distance from the goal position, where the pendulum is upright. 
An exponential reward is again employed.

For the inverted double pendulum task we use the \verb|InvertedDoublePendulum-v2|. It is similar to \verb|InvertedPendulum-v2|, except the pendulum now consists of two links. We only apply force to the cart  and have to stabilise the system to the upright positions. We add a wrapper to the default environment that is not modifying its behaviour, but changes slightly the interface, replacing a state variable corresponding to an angle with its sine and cosine values. This corresponds to an existing functionality from $\mathsf{PILCO}$ \cite{pilco} (see also \cite{pil_thesis}).

In the \verb|Swimmer-v2| from the OpenAI gym, a robot with two joints navigates a 2-d plane by "swimming" in a viscous fluid. Each joints are controlled by an actuator, and the system is rewarded for moving in the direction of the x-axis. This is a more challenging task, with an 8D state space, 2D control space, and nonlinear dynamics. 
Furthermore, it requires coordination between the two controllers for the robot to start moving towards the right direction: this makes the acquisition of a reward signal at the early stages of training hard \cite{guided_journal}.
The rewards are given for distance travelled in the positive direction of the x-axis, based on the position of the root link (to the right of Figure \ref{fig:swimmer}). 
This position variable however is not one of the state variables directly observed and, for this task, there is no specific goal position. 
Thus we define a linear reward for $\mathsf{PILCO}$, based on the x-axis velocity of the agent. Also (even in the plain $\mathsf{PILCO}$ version) we lightly penalise extreme angles at the joints, which leads to a smoother gait, allowing the policy to generalise outside of the planning horizon. 

\begin{table}[t]
    \parbox{0.53\linewidth}{
	\centering
	\resizebox{\linewidth}{!}{%
		\begin{tabular}{|c c c|} \hline
			\raggedleft
			Notation & Description                  & Python Var.  \\ \hline
			-		 &initial random rollouts        & J    \\
			N        & \# training episodes per run & N    \\ 
			-		 & Type of Controller           & Linear/RBF  \\
			dt		 & sampling period              & SUBS\\
			H		 & time steps per episode       & H    \\
			$\mu_0$  & initial state mean           & $m_{init}$   \\
			$\Sigma_0$ & initial state variance     & $S_{init}$    \\
			maxiter  & optimiser iterations         & maxiter    \\
			$\epsilon$ & max tolerable risk         &  th  \\
			\hline 
	\end{tabular}}
	\captionsetup{width=0.9\linewidth}
	\caption{\small{List of hyperparameters - notation, meaning and source-code variable}}
	\label{table:hypers}
    }
    \hfill
    \parbox{0.45\linewidth}{
	\centering
	\resizebox{\linewidth}{!}{%
		\begin{tabular}{|c c c c|} \hline
			& LinearCars & BAS & SafeSwimmer\\
			Con. Viol.  & 0.0$\pm$0.0 & 0.0$\pm$0.0 & 0.4$\pm$0.49\\ 
			Best return & -10.7$\pm$2.7 & 1.2$\pm$0.9& 11.6$\pm$8.2\\
			Max Episodes & 8    & 4 &  12\\
			Blocked Ep. & 1.4$\pm$1.5 & 0$\pm$0 & 1.1$\pm$1.3\\
			$\epsilon$       & 0.05 & 0.05 &   0.2  \\
			\hline 
	\end{tabular}}
	\captionsetup{width=0.9\linewidth}
	\caption{\small{$\mathsf{SafePILCO}$ results on constrained environments}}
	\label{table:safe_results}
	}
	\vspace{-0.4cm}
\end{table}

\subsection{Safe PILCO}

In this section we showcase the performance of the algorithm in environments with constraints
over the state space. 
The experimental structure is similar to that in Section \ref{sec:plain}, however we report metrics differently: we list the number of constraint violations incurred during training and the best performance achieved in episodes where the system has respected the constraints. 
Both these metrics are averaged over multiple random seeds, and the results are presented in Table \ref{table:safe_results}. 
We also report the average number of episodes (as blocked episodes) when the safety check has prohibited interaction with the system. 

\subsubsection{Linear Cars}
This scenario is similar to the one in \cite{polymenakos2019safe}, where two cars are approaching a junction, and the algorithm controls one of them by either braking or accelerating. 
The goal is for the controlled car to cross the junction as soon as possible, without causing a collision. The state space is 4-dimensional, with linear dynamics. 
The input $u$ has one dimension, proportional to the force applied to the first car.
To avoid a collision, the cars must not be simultaneously adjacent to the junction (set at the origin (0,0)). 
This can be formulated as a constraint of the form: $ |x^1 | > a  \; \mathrm{OR} \; | x^3 | > a$
over the position of the two cars. 
We want to encourage the first car to cross the junction as soon as possible: a simple reward could be $-1$ for every time-step where first car hasn't crossed the junction, and $+1$ otherwise. However, this is discontinuous and cannot be used by $\mathsf{SafePILCO}$ directly, so we use instead a linear reward, proportional to the position of the first car.

\subsubsection{Building Automation Systems}
We consider a problem in the domain of \emph{building automation systems}. 
The environment is given by \cite{cauchi2018benchmarks} (Case Study 2), which has developed a simulator
\footnote{Code for the BAS simulator:  \url{https://gitlab.com/natchi92/BASBenchmarks}}
based on real measurement data.
The task is to control the temperature in two adjacent rooms from a common heated air supply. 
The original cost function is the quadratic error between the temperatures in each of the two rooms and corresponding reference temperatures.
For $\mathsf{SafePILCO}$ we use the standard exponential reward function.

\subsubsection{Safe Swimmer}
This case study is based on the \verb|Swimmer-v2| environment, but we add the following constraints:
we require that the angles at the two joints remain below a certain threshold (95 $\deg$). 
Constraints of this sort are common in robotics, since pushing the joints to the 
edge of their functional ranges can lead to accumulated damage to the joints, the motors, or the robot links. 

\subsection{Results}
The results in Figure \ref{fig:plainpilco} and Table \ref{table:safe_results} show that $\mathsf{SafePILCO}$ is flexible and allows to tackle a wide selection of RL
problems, with good performance and  data efficiency. 
For reference, in \cite{wang2019benchmarking} 
model-based methods are evaluated on gym tasks, with the low-data regime having 200k data points and the high-data one 2 million points, while for the Swimmer $\mathsf{SafePILCO}$ uses only 625 points from $\sim$3000 interactions steps (also see Figure \ref{fig:plainpilco}).
Interestingly, \cite{wang2019benchmarking} reports that $\mathsf{PILCO}$ has managed to solve only  easy tasks (mean return of $\sim$0 on the Swimmer). 

\begin{table}[t]
	\centering
	\small
	\resizebox{\linewidth}{!}{%
		\begin{tabular}{|c|ccccccc|} \hline
			\raggedleft
			Variable & \multicolumn{7}{c|}{Tasks}  \\ 
			\hline
			-		 & MountainCar & InvPend & PendSwing & DoublePend  & Swimmer   & SafeCars  &    BAS \\
			State dim&  2          &   4     &    3      &      6         &    8   &    4      &    7   \\
			Control dim& 1         &   1     &    1      &      1         &    2   &    1      &    1   \\
			J		  & 2          &   5       &  4        & 5              & 15   &    5      &    4   \\
			N         & 4          &   3       &  8        & 10             & 10   &    8      &    3   \\ 
			Controller Type & RBF  &  RBF    &  RBF      & RBF            &  RBF   &    RBF    &    Linear  \\
			Basis Functions& 25    &   5     &  30       & 40             &  40    &    40     &    -   \\
			SUBS	 & 5           & 1       & 3         &  1             &  5     &    1      &    -   \\
			H		 & 25          &30       & 40        &  40            &  15    &    25     &    48  \\
			$\Sigma_0$ & $0.1\mathbf{I}$& $0.1\mathbf{I} $  & $\text{0.01diag[1,5,1]}$  & $0.5\mathbf{I}$ & $0.005\mathbf{I}$ & $0.1\mathbf{I}$ & $0.2\mathbf{I}$\\
			maxiter  & 100         & 100     & 50 & 120 &             100         &    20 & 25 \\
			\hline 
	\end{tabular}
	}
	\captionsetup{width=\textwidth}
	\caption{\small{List of hyperparameter values employed in the experiments}}
	\label{table:hypers2}
	\vspace{-0.5cm}
\end{table}


\section{Extensions and Conclusions} 
The $\mathsf{SafePILCO}$ software tool is a framework for safe and data-efficient policy synthesis, which are key features of the dynamics of physical systems. 
We have evaluated the software performance in a variety of standard benchmarks, and we have released a modular, extensible open-source implementation for reproducibility and further use by the community. 
Possible extensions include learning the reward function, joint training using a number of predicted trajectories, and making planning more effective in highly uncertain settings (including partially  observed ones).

\bibliographystyle{splncs04}
\bibliography{cite}

\appendix
\section{Related Work}
The native $\mathsf{PILCO}$ algorithm \cite{pil_thesis,pilco} is a policy search  framework \cite{policy_search_robotics}, 
employing Gaussian processes (GPs) \cite{gpbook} to learn the model dynamics and to maximise data efficiency. Extensions have been proposed, taking constraints into account \cite{pilco2,polymenakos2019safe}. 
In \cite{pilco2}, the constraints are incorporated as negative rewards, discouraging the system from visiting certain parts of the state space. However, these rewards have to be hand-tuned to balance performance and safety. Instead in \cite{polymenakos2019safe} an automatic procedure is introduced to formally synthesise policies satisfying spatial constraints, whilst additionally retaining safety during training --  $\mathsf{SafePILCO}$ is a tool implementing this procedure. 

Bayesian optimisation has also been used to train policy parameters \cite{berk,berk_swarm,safe}, towards data efficiency and (specific notions of) safety or invariance. 
Such methods also employ GPs, but do not learn system dynamics, instead mapping parameters to the loss/reward directly. 
This framework limits the number of policy parameters that can be tuned effectively.

Other model-based RL approaches have been proposed recently, describing the system dynamics through probabilistic models based on ensembles of deep neural networks \cite{chua2018deep,vuong2019uncertainty}, or Gaussian Processes \cite{chatzilygeroudis2017black,Vinogradska2018}. 
These methods provide viable alternatives to $\mathsf{PILCO}$, but have not been used in combination with safety requirements encoded as spatial constraints. Other model-based approaches with GPs either focus on stability \cite{for_mod} or take an approach \cite{koller2018learning} that provides more conservative guarantees, but which  restricts scalability by significantly increasing computational demands. 
Our method is the only one that maintains $\mathsf{PILCO}$'s favourable analytic properties and combines them with explicit state-space constraints for safe learning and safety goals. 

Finally, standard benchmarks are openly developed and maintained in the RL community \cite{openai,duan2016benchmarking,todorov2012mujoco}.
Our tool leverages some of these benchmarks and aims to provide an efficient and easy-to-use software package.

\section{Background and Theory}
\subsection{Problem formulation}
Our goal is to design a controller for an unknown system with general dynamics, 
which attains optimal performance in terms of accumulated reward. 
Further, the system trajectories must avoid unsafe regions of the state space. We assume the following:
\begin{itemize}
    \item a state space $X\subset \mathbb{R}^n$;  
    \item an input space $U \subset \mathbb{R}^m$; 
    \item a dynamical system with a transition function $x_{t+1}= f(x_t,u_t) + v_N$,
    where $v_N$ is assumed to be an i.i.d. Gaussian noise; 
    \item a reward function $r : X  \rightarrow \mathbb{R}$;
    \item a set $S \subset X$ of safe states and a corresponding set $D = X \setminus S$ of unsafe (dangerous) states; 
    we assume that these constraints $S$ are axis-aligned rectangles. 
\end{itemize}

Our task is to design a policy, $ \pi^\theta : X \rightarrow U $, with parameters $\theta$, 
which maximises the expected total reward over an episode while keeping the probability of violating the constraints low (e.g. lower than a given threshold).
We further require a minimal number of constraint violations \emph{during} training.
Since the system is initially unknown, and we are taking a data-driven approach, we cannot certify perfect compliance to all safety constraints without making further assumptions \cite{koller2018learning}.
Instead, we require the probability of the system to lie in safe states to exceed a threshold $\epsilon > 0$.
The sequence of states the system visits is referred to as its  \begin{it}trajectory\end{it} $ \mathbf{x} = \left(x_1,...,x_T\right)$.
We require all $x_i \in \mathbf{x}$ to be safe, implying that $x_i \in S$.  
We define an episode as a sequence of $T$ transitions, or $T$ time-steps, from some initial state $x_1$.

\subsection{Main algorithmic steps}
In the following we provide a succinct presentation of the main algorithmic steps. For more details see \cite{pil_thesis,pilco,polymenakos2019safe}.

\subsubsection{Model fit to data} 
We model the transition function with a Gaussian Process (GP) \cite{gpbook} that can provide predictions for a state $x_t$, given the previous state $x_{t-1}$ and the control input $u_{t-1}$. At any point the transition function PDF is a Gaussian distribution such that 
$
p(x_t|x_{t-1}, u_{t-1}) = \mathcal{N}(\mu_{t+1}, \Sigma_{t+1}),
$
where $\mu_{t+1}$ and $\Sigma_{t+1}$ are  predictive mean and (co)variance,  respectively. 
Our approach uses a squared exponential kernel with Automatic Relevance Determination \cite{gpbook}. The kernel hyperparameters are trained based on maximum marginal likelihood. 


\subsubsection{Long-term trajectory predictions} \label{sec:long}
Whilst single-step predictions of $x_{t+1}$ from $x_{t}$ are readily provided by standard GP methods, we wish to predict a sequence of states, $x_1, \dots, x_T$, over an episode. 
These predictions are performed iteratively, using the previous posterior prediction from the GP model as prior for the next. However, since the inputs to the GP are now themselves distributions, the subsequent predictive distribution needs to be approximated. This is performed by the computationally tractable means of Gaussian moment matching \cite{pil_thesis,girard2003gaussian}.
Hence, if $x_t \sim \mathcal{N}(\mu_{t}, \Sigma_{t})$, we write
$$
p(x_{t+1} | \mu_t, \Sigma_t, \theta) \approx  \mathcal{N}(\mu_{t+1}, \Sigma_{t+1}),
$$
in which we have dropped the dependency on $u_t$, since the policy is deterministic and $u_t$ is only dependent on $x_t$ and $\theta$.

\subsubsection{Evaluation of expected return and of safety} \label{sec:eval}
We combine the model-based prediction for the system trajectory with the reward function, which we assume to have the following form:
\begin{equation}
\label{eq:rew}
r(x) = \exp(- | x - x_{\text{target}} |^2 / \sigma_r^2),
\end{equation}
where $\sigma_r^2$ controls how rapidly the reward decays. Such a reward aims at driving (and possibly stabilising) the system towards a target state $x_{\text{target}}$.  
Since the state prediction at any time step is approximated by a Gaussian,
the expected reward can be written as:
\begin{equation}
    \mathbb{E}_{x_t}[r(x_t)]  = \int_S r(x) \mathcal{N}(\mu_t , \Sigma_t) dx,
\end{equation}
which can be calculated analytically, since both the reward function and the predictive distribution over the states belong to the exponential family.
We can express the expected return over an episode as:
\begin{equation}
    R^\pi(\theta) = \sum_{t=1}^T\mathbb{E}_{x_t}[r(x_t)].
\end{equation}
\sloppy
We note that each $x_t$ is distributed according to $p(x_t | \mu_{t-1}, \Sigma_{t-1}, \theta)$, hence the return depends on the policy $\pi^\theta$ and its parameters $\theta$.

Similarly, introducing $Q^\pi$ as the probability of the system being in safe states throughout the episode and under policy $\pi$, we can write:
\begin{align}
    & Q^\pi(\theta) \approx 
    \int_S ... \int_S p(x_1) p(x_2 | \mu_1, \Sigma_1, \theta) ... p(x_T | \mu_{T-1}, \Sigma_{T-1}, \theta) dx_1 ... dx_T, 
\end{align}
thus:
\begin{equation}
 Q^\pi(\theta) \approx \prod_{t=1}^T \int_S \mathcal{N}(\mu_t, \Sigma_t) dx_t = \prod_{t=1}^T q(x_t),
\end{equation}
where $q(x_t)$ is the probability of the system being in the safe parts of the state at time step $t$: 
\begin{equation} \label{qs}
    q(x_t) = \int_S \mathcal{N}(\mu_t , \Sigma_t) dx.
\end{equation}
Safety estimation thus reduces to the calculation of the probability mass of the Gaussian-distributed prediction at every state $x_t$, marginalised over the safe set $S$.
As anticipated earlier, 
we consider constraints that define an upper and/or lower bound for some of the state dimensions, as in $lb^d \leq x^d \leq ub^d$. The constraints can also come as conjunctive or disjunctive pairs, where for the system to be safe we require $x^{d_1} \in C^1 \text{ AND } x^{d_2} \in C^2$, or $x^{d_1} \in C^1 \text{ OR } x^{d_2} \in C^2$. 
For simplicity, let us assume that there are independent upper and lower bounds for each dimension, then we can write: 
$$
q(x_t) = \int_{lb^1}^{ub^1} \dots \int_{lb^D}^{ub^D}  p(x_t) dx_t^1 \dots dx_t^D,
$$
where $D$ is the dimension of the state $x_t$.
Numerical approaches to compute this quantity are proposed in \cite{genz1992numerical}. 

\subsubsection{Policy improvement}
After evaluating the current policy, the algorithm proposes a new candidate policy for evaluation.
The new policy can have a higher probability of respecting the constraints (making the policy safer) or an increase in the expected return.
As a secondary reward, evaluating and promoting safety, we use a scaled version of the probability of respecting the constraints throughout the episode.

The composite objective function, capturing both safety and performance (a risk-sensitive criterion according to \cite{safe_survey}) is defined ($\xi$ is a hyperparameter) as: 
\begin{equation}
    J^\pi(\theta) = R^\pi(\theta) + \xi Q^\pi(\theta). 
\end{equation}
The specific design choices made in the model formulation, for the reward function and for the safety constraints, allow one to \emph{analytically} calculate the gradient of the objective function with respect to $\theta$, as: 
$$ 
\frac{d J^\pi(\theta)}{d \theta} = \frac{d R^\pi(\theta)}{d \theta} + \xi \frac{d Q^\pi(\theta)}{d \theta}.
$$
We refer the reader to \cite{pilco,polymenakos2019safe} for more detailed derivations. 
We can then use any gradient-based optimiser to seek an optimal policy, according to the current model of the system dynamics. In particular, we employ L-BFGS-B \cite{byrd1995limited}, as implemented in SciPy. 

\subsubsection{Safety check and iteration}
Having obtained a new candidate policy, which is to be implemented on the real system, we wish to first verify that it is safe enough, 
as it is possible for an unsafe policy to be optimal in terms of $J$ (depending on the relative magnitudes of $R$ and $Q$).
We thus add the following \emph{safety check}: 
whenever the risk associated with the policy is higher than a predetermined \emph{risk threshold}, it prohibits implementation of the policy; 
it increases $\xi$ by a multiplicative constant;  
and it restarts the policy optimisation process. 
Further, we check whether the policy is too conservative. 
If the policy risk is significantly lower than the risk threshold, we implement the policy but also decrease $\xi$, 
allowing for a more performance-oriented optimisation at the next iteration of the algorithm.  

When the policy is implemented, we record new data from the real system. If the task is performed successfully, the algorithm terminates. If not, we use the newly available data to update the model, and to then repeat the overall process.  
This adaptive tuning of the hyperparameter $\xi$ guarantees that only safe policies
(according to the current GP model of the system dynamics) are implemented,
and overcomes the otherwise hard choice of a good initial value of $\xi$. 

\section{User Advice}  \label{sec:hypers}
\subsection{Setting hyperparameter values and troubleshooting}
In this section we provide practical advice to users who wish to solve new tasks after successfully installing the package and working through the examples provided. 
There are several hyperparameters that need to be set in advance, but these (in general) are related to aspects of the problem at hand: early experimentation can thus help to avoid exhaustive hyperparameter searches.  
We organise the rest of this section around the major components of the framework and comment on hyperparameter settings, as well as possible troubleshooting for each major component in turn.

We note that although the model is a crucial component of the algorithm, the associated hyperparameters do not need to be set in advance: indeed, signal variance, signal noise, and length scales can all be optimised when training the model.
In a low data-regime, however, optimisation can result in extreme values which lead to numerical instabilities. 
To avoid these issues we recommend:
\begin{itemize}
\item setting the signal noise to a fixed value (as done in the examples),
\item putting priors can be used to regularise hyperparameter values, such as the Gamma priors on the lengthscale hyperparameters and signal variance,
\item increasing the amount of data collected before the first run of the algorithm
\item reducing the iterations of the optimisation runs.
\end{itemize}

The exponential reward function from (\ref{eq:rew}), used for most experiments has two hyperparameters, the \emph{target}, $x_{target}$ and the \emph{weights}, $\sigma^2_r$. 
The hyperparameter values are not estimated from training data, so careful prior selection is required.
The target value should be the goal state to which a successful policy should drive the system. 
Intuitively, the weight matrix, $\sigma^2_r$, defines how quickly the reward is reduced as the distance between the current state and the goal state increases.
While the weight matrix can, in general, be any symmetric positive definite matrix, in all our test cases we use diagonal matrices.
For diagonal weight matrices, each value dictates how quickly the reward decays for a corresponding state variable.
We recommend these weights should take reasonable values between two extremes:
\begin{itemize}
\item high-magnitude values make the reward decay faster and make exploration harder (the reward signal becomes sparser);
\item low-magnitude values can make the reward gradient uninformative, or very small in magnitude, slowing down learning.
\end{itemize}

Furthermore, the reward function needs to promote safety by taking the constraints into account. 
The weighting between the reward and constraint functions is automated. 
However, a mismatch between the risk calculated and the constraints we have imposed can point to a model failure, or to a mistaken constraint formalisation. RL agents are known to exploit the reward functions provided, an issue often referred to as \emph{reward hacking} or \emph{goal misalignment} \cite{amodei2016concrete,everitt2018towards}.
When that happens, the agent consistently collects high reward, but its behaviour is far from what a human would consider as solving the task.
In such cases, reconsidering the reward function hyperparameters (target and weight matrix) is advised.

Another issue we find can arise is that of learning being inhibited by a bad controller initialisation.  
While we try to resolve this issue automatically, it can still happen that the controller parameters are chosen so that the agent takes no actions whatsoever, or that the policy function ($u = \pi^\theta(x)$) is approximately constant in relevant regions of the state space visited, resulting in (near) zero gradients. 
In such scenarios, normalising the data (included in the functionality of the package) can often help.

More extensive advice on troubleshooting, with examples and code, can be found in a Jupyter notebook associated to the $\mathsf{SafePILCO}$ package.

\subsection{Algorithm assumptions and  restrictions} \label{sec:assum}

We delineate here some of the restrictions of the algorithm, 
for two purposes: firstly, for users to easily assess whether the current version fits their application, and secondly, to outline directions for future research.

A strong assumption of the $\mathsf{PILCO}$ algorithm is that the environment is fully observable and Markovian. 
That means that the full state can be observed by the agent at every time step, and that all information relevant for predicting the next state is captured by the current state measurement and selection of the present control input. 
Empirically we have seen that small amounts of noise in the state measurements, despite violating this assumption, can be beneficial, improving numerical stability, but performance deteriorates as the magnitude of the noise increases (resulting in actual lack of observability). 
Since extensions in the direction of partial observations exist for the original $\mathsf{PILCO}$ algorithm \cite{mcallister2016data}, we expect they might as well apply to this setup, and would constitute a relevant extension of this project.  

The reward function is assumed to be predefined, in closed form, so that for a Gaussian distributed state, the expectation of the reward can be efficiently calculated. 
Reward shaping can mitigate this issue, however it is out of the scope of this paper. 
In future work, the reward function can be approximated with a Gaussian Mixture Model, which would maintain Gaussian features for noisy states.
Learning the reward function from observations has been used in other model-based RL approaches \cite{franccois2019combined}.

Our model is based on GPs with squared exponential kernels, which are underpinned by an assumption of universal smoothness and differentiability of the system dynamics. 
This assumption does not necessarily hold for all environments, 
e.g. whenever contact dynamics need to be modelled, or under hybrid/switching dynamics. 
This is a significant challenge for the kind of model we propose, hence replacing this component would require rethinking and replacing the moment matching approximation used for multi-step planning. Work in this direction \cite{Vinogradska2018}, replacing moment matching by numerical quadrature, can be a promising approach. 

A final consideration is that the planning step is based on Gaussian-distributed predictions. This assumption can be limiting in several cases, particularly when the task at hand has high initial uncertainty, e.g. when each episode starts from an arbitrary state. Then, the unimodal, Gaussian-distributed trajectory prediction is uninformative, and $\mathsf{PILCO}$ often fails to estimate useful gradients of the objective function with respect to the policy parameters. Training from multiple distinct initial states can help to mitigate this issue \cite{pil_thesis}. 

\end{document}